\documentclass[11pt]{article}

\usepackage[final]{acl}

\usepackage{times}
\usepackage{latexsym}
\usepackage{subcaption}
\usepackage{amsmath}
\usepackage{booktabs}

\usepackage[T1]{fontenc}
\usepackage[utf8]{inputenc}

\usepackage{microtype}

\usepackage{inconsolata}

\usepackage{graphicx}

\title{Capacity Constraints and the Multilingual Penalty for Lexical Disambiguation}

\author{
  Sean Trott
  \\
  Rutgers University - Newark
  \\
  \texttt{sean.trott@rutgers.edu}
  \And
  Pamela D. Rivi\`ere
  \\
  Rutgers University - Newark
  \\
  \texttt{pamela.riviereruiz@rutgers.edu}
}

\begin{document}
\maketitle
\begin{abstract}
Multilingual language models (LMs) sometimes under-perform their monolingual counterparts, possibly due to \textit{capacity limitations}. We quantify this ``multilingual penalty'' for lexical disambiguation---a task requiring precise semantic representations and contextualization mechanisms---using controlled datasets of human relatedness judgments for ambiguous words in both English and Spanish. Comparing monolingual and multilingual LMs from the same families, we find consistently reduced performance in multilingual LMs. We then explore three potential capacity constraints: representational (reduced embedding isotropy), attentional (reduced attention to disambiguating cues), and vocabulary-related (increased multi-token segmentation). Multilingual LMs show some evidence of all three limitations; moreover, these factors \textit{statistically account for} the variance formerly attributed to a model's multilingual status. These findings suggest both that multilingual LMs do suffer from multiple capacity constraints, and that these constraints correlate with reduced disambiguation performance. 
\end{abstract}

\section{Introduction}

In principle, training language models (LMs) on multiple languages should facilitate efficient cross-linguistic generalizations and widespread practical deployment. Yet as the number of languages on which a model is trained increases, multilingual models sometimes under-perform their monolingual counterparts \citep{conneau-etal-2020-unsupervised, chang-etal-2024-multilinguality, wang-etal-2020-negative, pfeiffer-etal-2022-lifting, blevins-etal-2024-breaking}, possibly due to \textit{insufficient model capacity} \citep{chang-etal-2024-multilinguality}. Here, we quantify the multilingual penalty associated with \textit{lexical disambiguation}, and explore several distinct routes by which reduced capacity might manifest in multilingual LMs. We focus on lexical disambiguation in particular because it demands both precise semantic representations and well-developed mechanisms for disambiguation in context \citep{riviere2025startmakingsensesdevelopmental}; additionally, it has been extensively characterized across multiple disciplines, including NLP \citep{schlechtweg-etal-2018-diachronic, schlechtweg2025comedi, haber-poesio-2021-patterns-polysemy, trott-bergen-2021-raw}.

One potential source of reduced capacity is \textit{representational}: notwithstanding cross-linguistic generalizations, multilingual LMs must compress more linguistic knowledge into the same number of dimensions, potentially reducing \textit{within-language isotropy} \citep{ethayarajh-2019-contextual}. Because some multilingual LMs maintain language-specific representations \citep{chang-etal-2022-geometry}, embeddings for words from a given language might occupy a narrower cone of vector-space (i.e., \textit{anisotropy}) than in a monolingual model. While the downstream consequences of anisotropy are still under debate  \citep{machina-mercer-2024-anisotropy, godey-etal-2024-anisotropy, rajaee-pilehvar-2022-isotropy, rudman-etal-2022-isoscore}, reduced isotropy could limit a model's ability to discriminate distinct word meanings in context  (e.g., ``tree \textit{bark}'' vs. ``dog \textit{bark}''). 

A related possibility is reduced \textit{attentional} capacity. Attention heads likely play some role in contextualizing the meaning of target ambiguous words \citep{riviere2025startmakingsensesdevelopmental}. It is implausible that the same attention head could perform disambiguation across diverse languages---yet it is equally implausible that specialized attention heads could develop for each individual language. On net, this could limit a multilingual LM's ability to contextualize ambiguous words using attention mechanisms. 

A third possibility is reduced \textit{vocabulary} capacity. Multilingual LMs must cover far more words with a comparably sized vocabulary; consequently, more words will be segmented into multiple \textit{sub-word tokens}, which in turn may not reflect meaningful morpheme boundaries \citep{arnett-etal-2024-different}. This could be particularly problematic for disambiguation: when target words are split across tokens, extracting a coherent representation might be less straightforward. Indeed, there is some evidence that impaired performance in multilingual LMs can be attributed to challenges with tokenization \citep{rust-etal-2021-good}.

In the current work, we evaluate disambiguation in both English and Spanish, using tightly controlled datasets of human judgments about ambiguous words in minimal pair contexts \citep{trott-bergen-2021-raw, riviere-etal-2025-evaluating}. We also use monolingual/multilingual ``minimal pairs'', i.e., from the same model family (e.g., BERT). In Section \ref{subsec:penalty}, we first quantify the multilingual penalty; we then quantify and explore the potential \textit{correlates} of this penalty and ask which best accounts for the reduction in disambiguation performance (Section \ref{subsec:factors}). All data and code required to reproduce the analyses in the current manuscript is available on GitHub: \url{https://github.com/seantrott/multilingual_disambiguation}.

\section{Current Work}

\subsection{Methods}\label{subsec:methods}

\subsubsection{Datasets}

We used two datasets containing human relatedness judgments about ambiguous words across minimal pair contexts (e.g., ``She liked the marinated lamb'' vs. ``She liked the friendly lamb''). RAW-C (Relatedness of Ambiguous Words---in Context) contained judgments for 672 English sentence pairs \citep{trott-bergen-2021-raw}, while SAW-C (Spanish Ambiguous Words---in Context) contained judgments for 812 Spanish sentence pairs \citep{riviere-etal-2025-evaluating}. These datasets were selected for their tight experiment control (ambiguous words were embedded in minimal pair contexts) and their \textit{graded} human judgments about relatedness (consistent with recent work arguing for continuous representations of word meaning \citep{elman2009meaning, li2021word, trott2023large, li2024semantic}). Both datasets had been anonymized by previous work.

\subsubsection{Models}

We assessed 24 unique model instances: 10 monolingual English models, 10 monolingual Spanish models, and 4 multilingual models. Our selection of models was guided by two primary factors: first, because the Spanish disambiguating cue occurred \textit{after} the target word, we identified a range of \textit{bidirectional models} \citep{riviere-etal-2025-evaluating}; and second, we selected model families that contained both monolingual English models and monolingual Spanish models, as well as multilingual variants. Model families included ALBERT/ALBETO \citep{DBLP:journals/corr/abs-1909-11942, CaneteCFP2020}, BERT/BETO \citep{DBLP:journals/corr/abs-1810-04805, CaneteCFP2020}, DistilBERT/DistilBETO \citep{sanh2019distilbert, CaneteCFP2020}, and RoBERTa \citep{DBLP:journals/corr/abs-1907-11692, gutierrez2022maria}; all but one of these families (ALBERT) also included at least one multilingual variant, e.g., XLM-RoBERTa \citep{conneau-etal-2020-unsupervised}, mBERT \citep{DBLP:journals/corr/abs-1810-04805}, and DistilBERT \citep{sanh2019distilbert}. A table listing all individual model instances (along with their number of parameters) can be found in Appendix \ref{sec:app_list}.

\subsection{Results}\label{subsec:results}

\subsubsection{Quantifying the ``Multilingual Penalty''}\label{subsec:penalty}

As in past work \citep{trott-bergen-2021-raw, riviere-etal-2025-evaluating}, disambiguation performance was assessed by presenting each sentence pair from each dataset to a given model instance. We then calculated the cosine distance between the contextualized embeddings of the target word (e.g., ``lamb'') across all layers of that model. Finally, we regressed human relatedness judgments against cosine distance and used the resulting $R^2$ as a index of how successfully representations from that layer predicted relatedness. Multilingual models were evaluated using both datasets, and monolingual models were evaluated using only the dataset in the target language.

As depicted in Figure \ref{fig:metrics}, models ranged considerably in their maximum performance, though none surpassed human inter-annotator agreement \citep{trott-bergen-2021-raw, riviere-etal-2025-evaluating}. We then estimated the ``multilingual penalty'' by regressing disambiguation performance from each layer of each model models ($R^2$) against several factors: Log Parameter Count, Language (English vs. Spanish), and Multilingual status (Yes vs. No); we also included random intercepts for each model. Performance improved for bigger models $[\beta = 0.09, SE = 0.03, p = 0.001]$ and later layer depths $[\beta = 0.2, SE = 0.01, p < .001]$; multilingual models exhibited reduced performance even controlling for these other factors $[\beta = -0.16, SE = 0.04, p < .001]$. That is, representations from multilingual LMs were worse at predicting relatedness than were representations from equivalent layers of monolingual LMs. (Note that equivalent results were obtained using only maximal $R^2$ instead of the layer-wise measure; see Appendix \ref{sec:app_max_r2}).

\begin{figure}
    \centering
    \includegraphics[width=.9\linewidth]{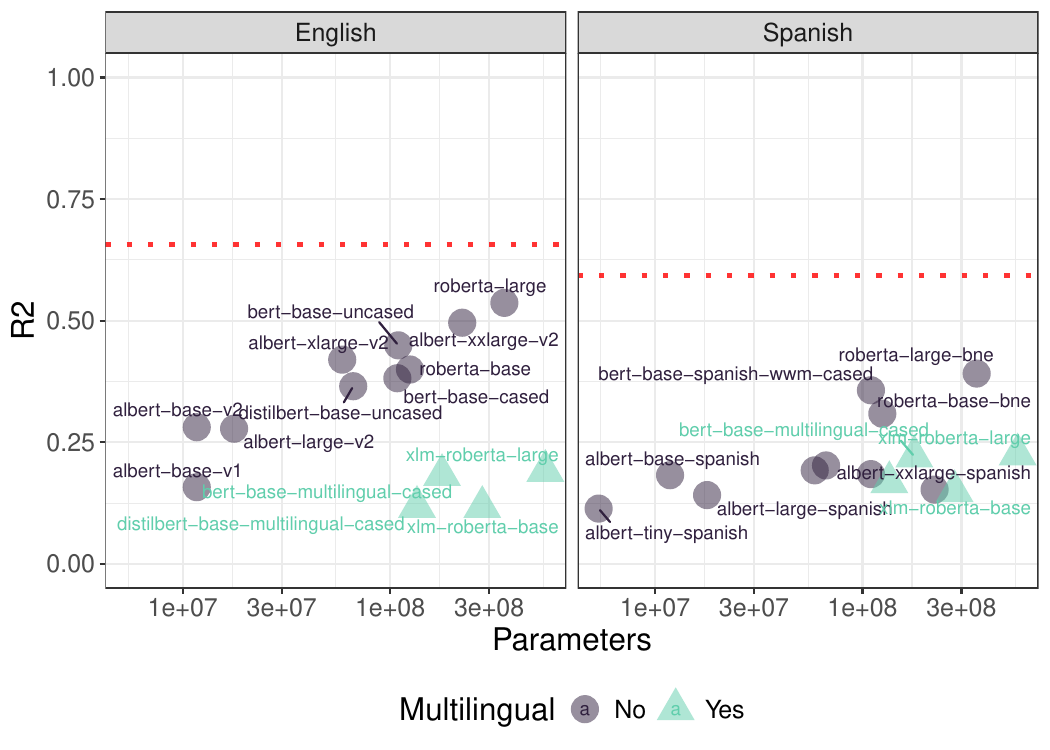}
    \caption{The best-performing layers of multilingual models generally exhibited reduced performance (as measured by $R^2$) compared to monolingual models of equivalent size.}
    \label{fig:r2_curse}
\end{figure}

\subsubsection{Decreased Representational Isotropy in Multilingual Models}

\begin{figure*}
    \centering
    \begin{subfigure}[t]{0.48\textwidth}
        \centering
        \includegraphics[width=\textwidth]{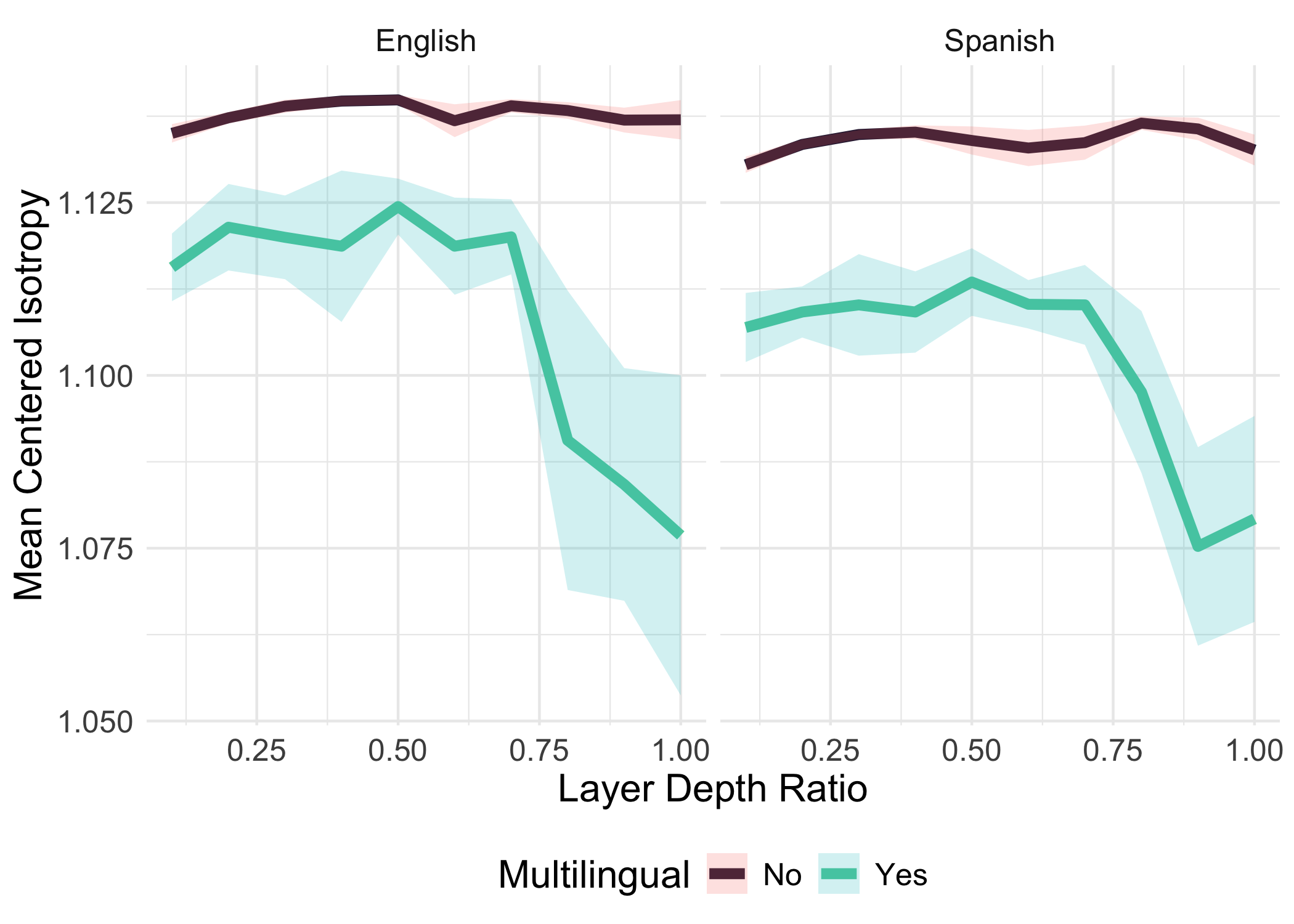}
        \caption{Multilingual models exhibited reduced embedding isotropy at equivalent layer depths.}
        \label{fig:isotropy}
    \end{subfigure}
    \hfill
    \begin{subfigure}[t]{0.48\textwidth}
        \centering
        \includegraphics[width=0.9\linewidth]{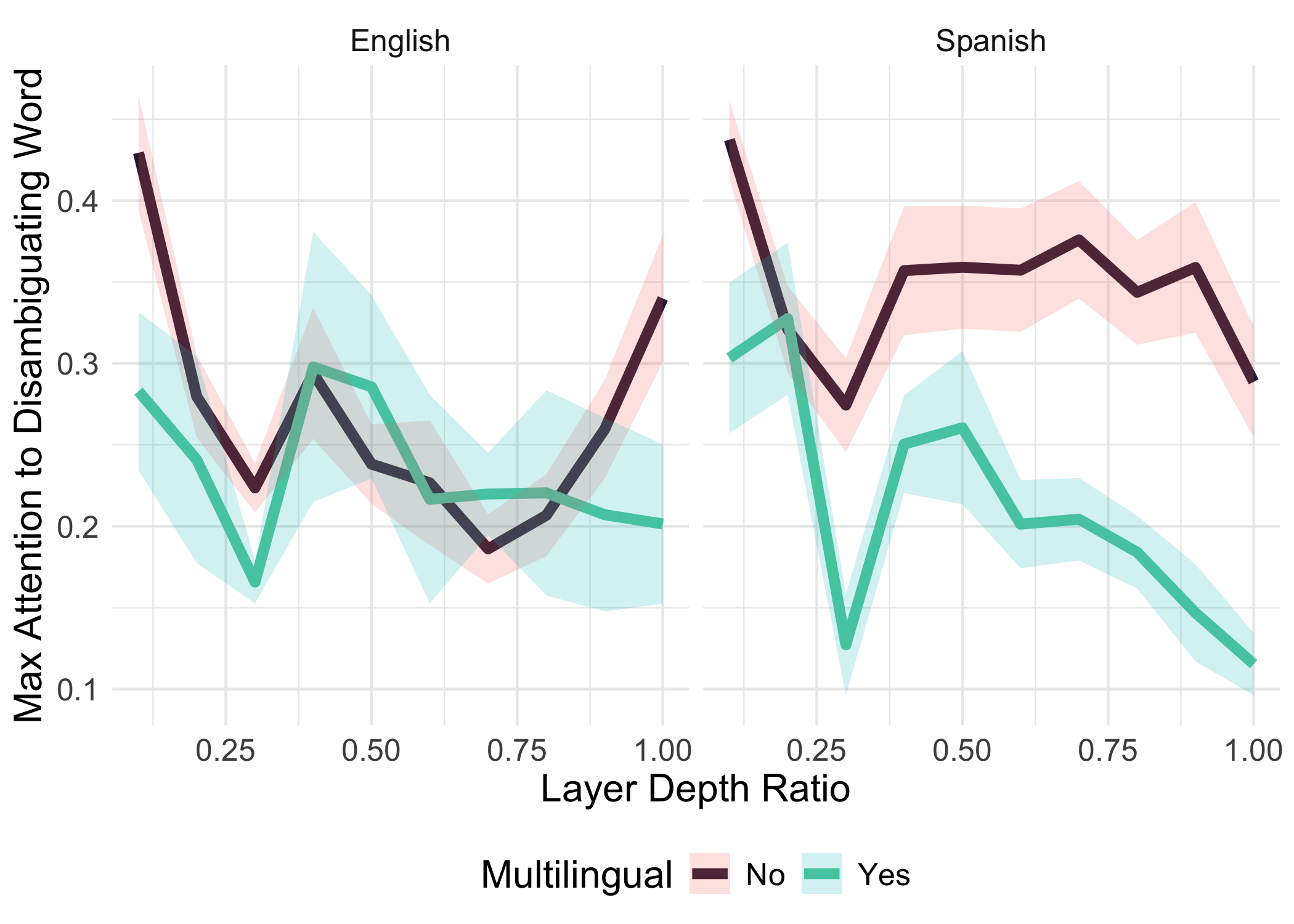}
        \caption{Attention heads in multilingual models directed less attention towards disambiguating cues in Spanish sentences than those in monolingual models.}
        \label{fig:attn}
    \end{subfigure}
    \caption{Compared to their monolingual counterparts, multilingual models models showed evidence of reduced isotropy (left) and somewhat reduced attention to disambiguating cues (right).}
    \label{fig:metrics}
\end{figure*}

As discussed above, there is no universally agreed-upon metric for evaluating the degree of \textit{isotropy} in LM embeddings \citep{ethayarajh-2019-contextual, rudman-etal-2022-isoscore}. We focus here on Centered Isotropy, or CI, which is calculated by first centering and normalizing all embeddings for a sentence, then computing the average cosine distance between each pair of embeddings (see Appendix \ref{app:iso_metrics} for alternative metrics). For each layer of each model, we calculated the mean CI for each input sentence. We then fit a linear mixed model with mean CI as the dependent variable, fixed effects of Layer Depth, Multilingual status (and its interaction with Layer Depth), Language, and Log Parameter Count; and random intercepts for Target Word and LM. Multilingual LMs were associated with reduced mean CI overall $[\beta = -0.02, SE = 0.004, p < 0.001]$ (see also Figure \ref{fig:isotropy}); moreover, this reduction was exacerbated at later layer depths $[\beta = -0.01, p < .001]$.

\subsubsection{Decreased Attention to Disambiguating Cues in Multilingual Models}

Disambiguation performance is likely linked to the degree of \textit{attention} directed from an ambiguous word (e.g., ``lamb'') to potential disambiguating cues to the disambiguating cue (e.g., ``marinated'') \citep{riviere2025startmakingsensesdevelopmental}. For each attention head in each model, we calculated the attention score between the target word and the disambiguating cue. We then aggregated these scores by layer, computing both the average and maximum attention across all heads in a layer.

In the English dataset (RAW-C), we found no evidence that multilingual models exhibited reduced attention to disambiguating cues (for either metric). However, we did observe differences in maximal attention to disambiguating cues in the Spanish dataset (SAW-C), particularly in later layers of multilingual models (see Figure \ref{fig:attn}). This was corroborated by the results of a statistical analysis regressing Max. Attention against Layer Depth, Multilingual status, Log Parameter Count, Language, an interaction between Language and Multilingual status, and random intercepts for model. The interaction effect was significant, with reduced attention for multilingual models tested in Spanish $\beta = 0.09, SE = 0.03, p = .01]$. (Note that the interaction was also significant, albeit smaller, when predicting mean attention.)

\subsubsection{Increased Rate of Multi-Token Words for Multilingual Models}

For both datasets, we counted the number of tokens corresponding to both the target word (e.g., ``lamb'') and disambiguating cue (e.g., ``marinated'') for each LM's tokenizer. We then built a series of linear mixed models predicting Number of Tokens (for the Target and Disambiguating Cue), with Multilingual status, Language, and Log Parameter Count as fixed effects, and random intercepts for Model, Target Word, and Sentence. As predicted, multilingual status was consistently associated with a higher number of tokens for both the target word $[\beta = 0.23, SE = 0.01, p < 0.001]$ and the disambiguating cue $[\beta = 0.43, SE = 0.04, p < .001]$; see also Figure \ref{fig:tokens}.

\subsubsection{Which Factors Account for Reduced Disambiguation?}\label{subsec:factors}

We then asked which factors accounted for the ``multilingual penalty''. We built a series of linear mixed models predicting layer-wise disambiguation performance from each factor---Cumulative Maximum Attention to the disambiguating cue across all layers up to layer $\ell$; mean CI from $\ell$; and the mean number of tokens in the target word---and assessed their fit using $AIC$ \citep{akaike2003new} (lower $AIC$ corresponds to better fit). All models included baseline covariates (Log Parameter Count, Layer Depth) and random intercepts for model; AIC values were rescaled to this baseline.

\begin{figure}
    \centering
    \includegraphics[width=.9\linewidth]{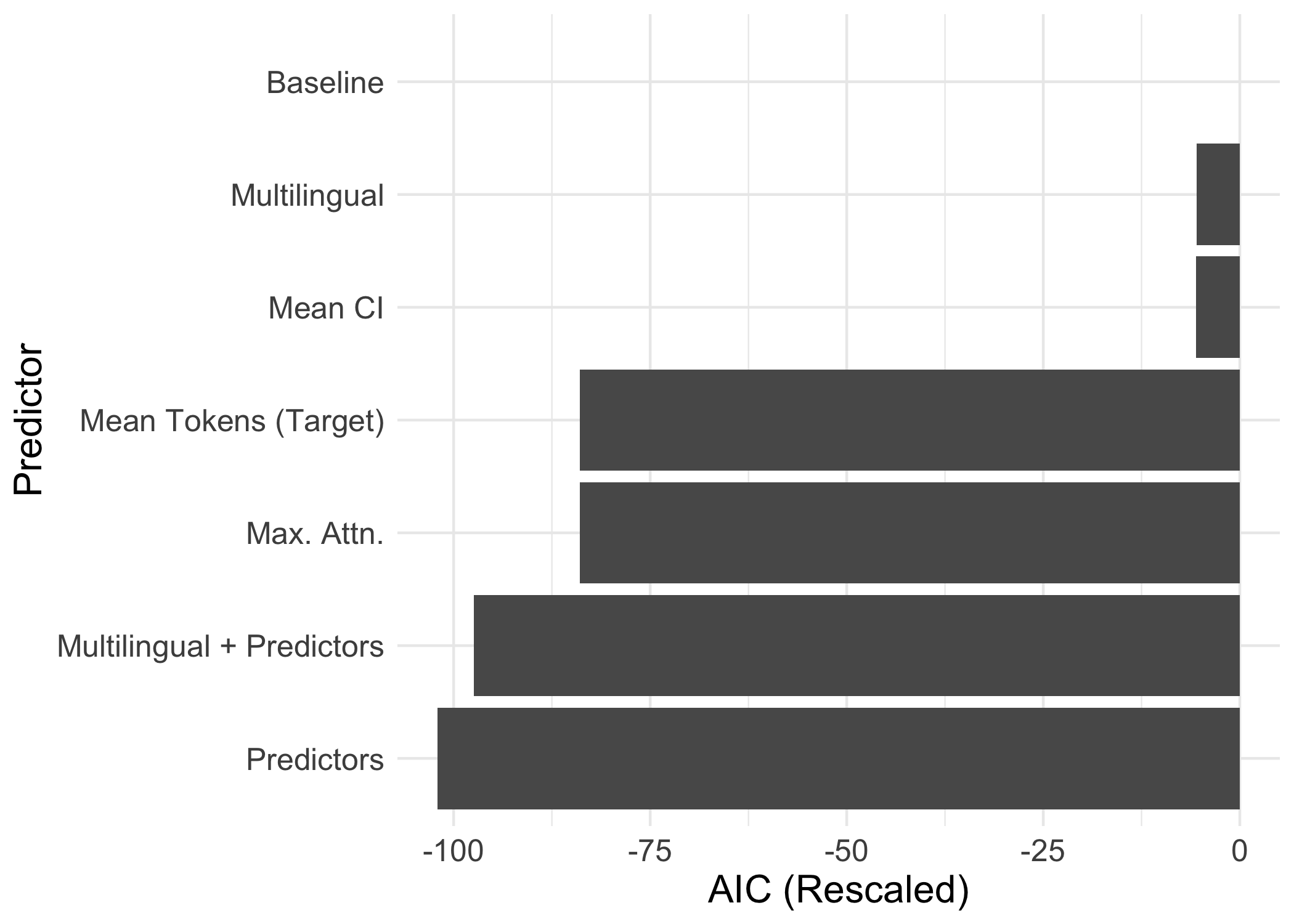}
    \caption{The $AIC$ (scaled to the Baseline model) associated with linear mixed models predicting disambiguation performance from various factors (lower is better).}
    \label{fig:factors}
\end{figure}

As depicted in Figure \ref{fig:factors}, mean CI represented only a marginal improvement over the Baseline model---roughly equivalent to Multilingual status alone. Tokenization and attention yielded substantially better fit ($\Delta AIC > 78$). Crucially, when all three factors were combined, adding Multilingual status \textit{hurt} model fit: the parameter was unnecessary because the other factors already captured its explanatory power. In the full model, increased isotropy predicted increased $R^2$ $[\beta = 1.2, SE = 0.27, p < .001]$ and multi-token words predicted decreased $R^2$ $[\beta = -0.58, SE = 0.13, p < .001]$. Counterintuitively, increased attention was associated with decreased performance $[\beta = -0.14, SE = 0.02, p < .001]$.

\section{Discussion}\label{sec:discussion}

We set out to quantify and explain the apparent penalty faced by multilingual LMs in disambiguation tasks. First, we confirmed that multilingual LMs consistently under-performed their monolingual counterparts on a disambiguation task in both English and Spanish (Section \ref{subsec:penalty}). Second, we found that multilingual LMs also displayed evidence of reduced isotropy (in both languages), reduced attentional capacity (in Spanish), and a higher rate of multi-token words (in both languages)---all potential \textit{correlates} of a multilingual penalty. Third, we found that the combination of these factors statistically accounted for the variance explained by an LM's multilingual status (Section \ref{subsec:factors}): that is, variance in isotropy, attention, and tokenization better accounted for variance in disambiguation performance than did a factor indicating whether an LM was multilingual. These results confirm that multilingual LMs do suffer from multiple kinds of \textit{capacity limitations}, consistent with prior work \citep{chang-etal-2024-multilinguality}; and moreover, that these \textit{correlate} with reductions in disambiguation performance. Although this work is limited in scope and inferential power (see Section \ref{sec:limitations}), it represents a proof-of-concept that at least in a subset of LMs, the multilingual penalty is correlated with measurable, relatively interpretable factors.

\section{Limitations}\label{sec:limitations}

A key limitation is \textit{scope}: although the datasets benefited from tight experimental control, they were limited in size and covered only two languages (English and Spanish). Similarly, we relied on LMs from a restricted set of families, none of which are considered state-of-the-art; this limitation was driven in part by the need to rely on bidirectional LMs (given that the ambiguous words in Spanish were always disambiguated by a word following the target), and by our aim to \textit{match} multilingual LMs with monolingual LMs with similar training protocols and architectures. However, future work could investigate whether this multilingual penalty is observed in larger, state-of-the-art multilingual LMs across a variety of languages---and whether the same explanatory factors (i.e., reduced isotropy, reduced attention to disambiguating cues, and more multi-token words) consistently co-vary with multilingual status.  

A related limitation is that our category of ``multilingual'' LM was quite coarse. To the extent that it exists, the multilingual penalty likely depends on the number and distribution of languages on which an LM is trained \citep{chang-etal-2024-multilinguality}. Future work could also assess LMs trained on a small number of related languages---and also vary the relative balance of training volume across languages---to investigate how much the multilingual penalty (and its correlates) depends on the amount and type of multilinguality.

Certain findings were also surprising and raise questions about interpretation. For instance, we found a significant \textit{negative} relationship between attention to disambiguating cues and overall disambiguation performance---the opposite of our predictions. One possibility is that better performance is actually driven by more distributed patterns of attention, perhaps reflecting greater redundancy; alternatively, our operationalization of ``attention to the disambiguating cue'' (i.e., the maximum attention from a given layer) could simply be flawed. This result also highlights the challenges of relying on attention maps in explanations of LM behavior \citep{jain-wallace-2019-attention, wiegreffe-pinter-2019-attention}, particularly in the absence of causal intervention or a fine-grained analysis of training dynamics \citep{riviere2025startmakingsensesdevelopmental}.

This leads to a final limitation: notably, our analyses were \textit{correlational} and do not establish a causal role for the factors we identified. It is possible that a single factor causally accounts for each of the other variables (e.g., perhaps differences in tokenization produce differences in measured isotropy or attention), or even that some unmeasured factor is truly responsible for the multilingual penalty. Future research could ask whether \textit{causally intervening} on these factors---i.e., isotropy, attention to disambiguating cues, and tokenization---improves disambiguation performance. 

\section{Ethical Considerations}

The primary ethical concerns relate to the inferential limitations discussed above: it is possible that limitations in our evaluation of multilingual LMs could lead to an underestimation (or overestimation) of their performance, which would have downstream effects on the relative risk of relying on such models.

\bibliography{custom}

\appendix

\section{Full list of models}\label{sec:app_list}

\begin{table}[htbp]
\centering
\small
\caption{Language Models by Family}
\begin{tabular}{lcc}
\toprule
\textbf{Model} & \textbf{Multi.} & \textbf{\# Params} \\
\midrule
bert-base-cased & No & $\sim$ 108M \\
bert-base-uncased & No & $\sim$ 109M \\
bert-base-spanish-wwm-cased & No & $\sim$ 110M \\
bert-base-spanish-wwm-uncased & No & $\sim$ 110M \\
bert-base-multilingual-cased & Yes & $\sim$ 178M \\
\midrule
distilbert-base-uncased & No & $\sim$ 66M \\
distilbert-base-spanish-uncased & No & $\sim$ 67M \\
distilbert-base-multilingual-cased & Yes & $\sim$ 135M \\
\midrule
albert-tiny-spanish & No & $\sim$ 5M \\
albert-base-v1 & No & $\sim$ 12M \\
albert-base-v2 & No & $\sim$ 12M \\
albert-base-spanish & No & $\sim$ 12M \\
albert-large-v2 & No & $\sim$ 18M \\
albert-large-spanish & No & $\sim$ 18M \\
albert-xlarge-v2 & No & $\sim$ 59M \\
albert-xlarge-spanish & No & $\sim$ 59M \\
albert-xxlarge-v2 & No & $\sim$ 223M \\
albert-xxlarge-spanish & No & $\sim$ 223M \\
\midrule
roberta-base & No & $\sim$ 125M \\
roberta-base-bne & No & $\sim$ 125M \\
roberta-large & No & $\sim$ 355M \\
roberta-large-bne & No & $\sim$ 355M \\
xlm-roberta-base & Yes & $\sim$ 278M \\
xlm-roberta-large & Yes & $\sim$ 560M \\
\bottomrule
\end{tabular}
\label{tab:models}
\end{table}

\section{Analysis of maximal $R^2$}\label{sec:app_max_r2}

We carried out an additional analysis investigating which factors predicted \textit{maximal} $R^2$. First, we replicated the analysis of the multilingual penalty. As in the main manuscript, we found that overall performance was higher for bigger models $[\beta = 0.15, SE = 0.03, p < .001]$ and lower for models tested in Spanish $[\beta = -0.1, SE = 0.03, p = .002]$. Crucially, multilingual models exhibited reduced performance even controlling for these other factors $[\beta = -0.22, SE = 0.04, p < .001]$. That is, an LM's multilingual status was associated with a $0.22$ \textit{decrease} in $R^2$ relative to models of an equivalent size tested on the same dataset.

\section{Additional Isotropy Metrics}\label{app:iso_metrics}

\begin{figure*}
    \centering
    \begin{subfigure}[t]{0.48\textwidth}
        \centering
        \includegraphics[width=\textwidth]{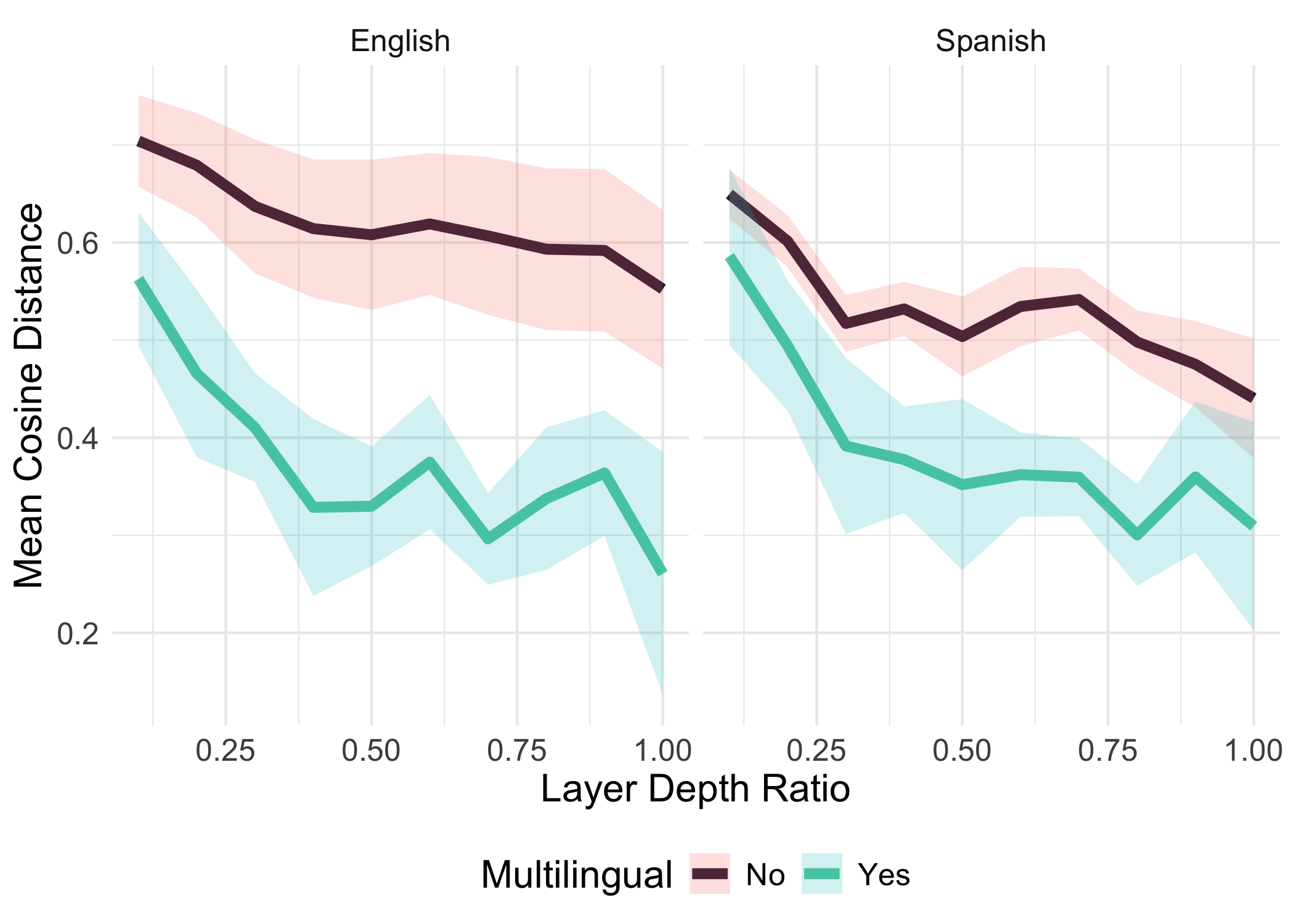}
        \caption{Embeddings from multilingual models were closer on average than embeddings from layers of an equivalent depth of monolingual models.}
        \label{fig:cd_isotropy}
    \end{subfigure}
    \hfill
    \begin{subfigure}[t]{0.48\textwidth}
        \centering
        \includegraphics[width=0.9\linewidth]{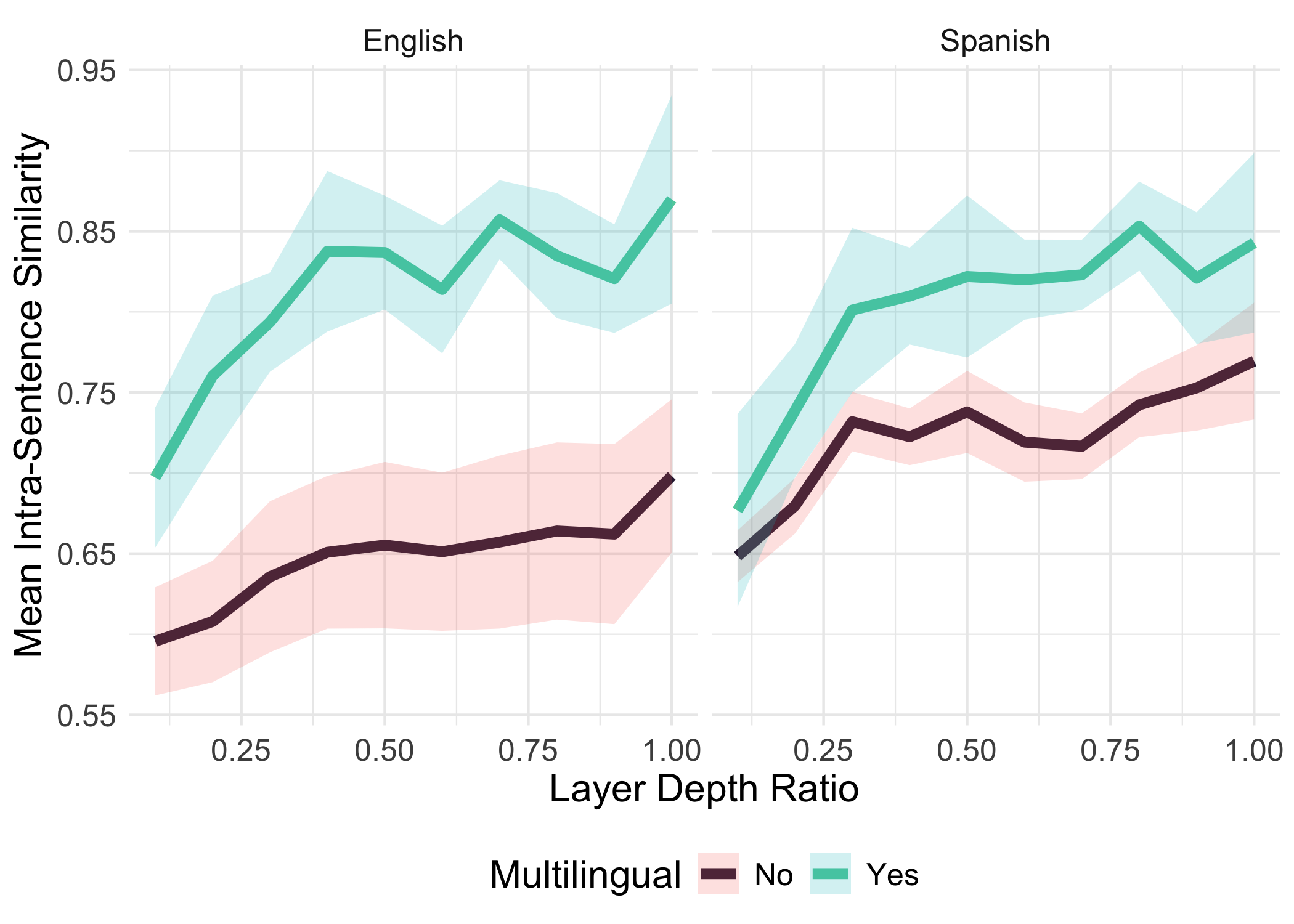}
        \caption{Individual token embeddings from multilingual models were more similar on average to the mean sentence embedding than embeddings from layers of an equivalent depth of monolingual models.}
        \label{fig:iss_isotropy}
    \end{subfigure}
    \caption{Multilingual models showed evidence of reduced isotropy (lower average cosine distance between token embeddings; and higher cosine similarity between individual token embeddings and the sentence average) relative to monolingual models. }
    \label{fig:metrics_appendix}
\end{figure*}

As noted in the primary manuscript, researchers use different metrics for evaluating embedding isotropy, which suffer from different advantages and disadvantages \citep{rudman-etal-2022-isoscore}. In addition to Mean Centered Isotropy, we evaluated Mean Cosine Distance (the average cosine distance between all token embeddings for a given sentence from a given layer) and Intra-Sentence Similarity (the cosine distance between each individual token embedding and the average embedding for the sentence from a given layer). In general, higher Mean Cosine Distance is interpreted as \textit{more} isotropic (i.e., embeddings are more widely dispersed), and higher Intra-Sentence Similarity is interpreted as \textit{less} isotropic (i.e., embeddings are all similar to the sentence average). As with Mean Centered Isotropy, we observed evidence of decreased isotropy in multilingual models compared to their monolingual counterparts, particularly at later layers, in both metrics (see Figure \ref{fig:metrics_appendix}).

\section{Increased Rate of Multi-Token Words}

As described in the primary manuscript, tokenizers for multilingual models were more likely to segment target words (and disambiguating words) into multiple tokens. A visual comparison of this difference is illustrated in Figure \ref{fig:tokens} below.

\begin{figure*}
    \centering
    \includegraphics[width=\linewidth]{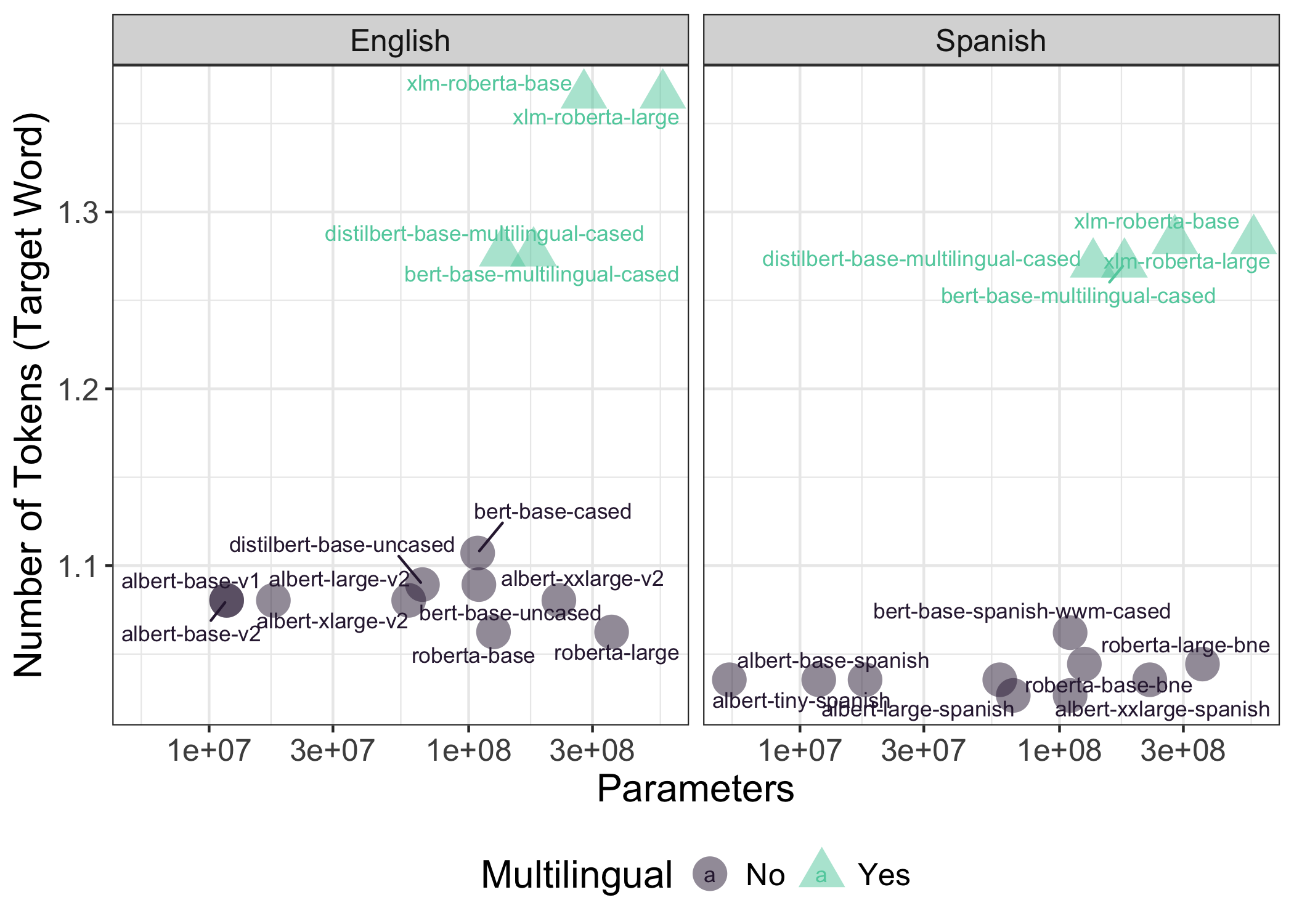}
    \caption{Multilingual models consistently segmented target words into more tokens than monolingual models.}
    \label{fig:tokens}
\end{figure*}

\end{document}